\documentclass[sigconf]{acmart}

\AtBeginDocument{%
  }

\setcopyright{acmlicensed}
\copyrightyear{2018}
\acmYear{2025}
\acmDOI{XXXXXXX.XXXXXXX}
\acmConference[MM'25]{ACM Conference}{October 27--31,
  2025}{Dublin, Ireland}
\acmISBN{978-1-4503-XXXX-X/2018/06}

\acmSubmissionID{5584}

\usepackage{subcaption}
\usepackage{svg}
\usepackage{url}
\usepackage{booktabs}
\usepackage{enumitem}
\usepackage{bbding}
\usepackage{pifont}
\usepackage{multirow} 
\usepackage{wasysym}
\usepackage{utfsym}
\usepackage{bm}
\usepackage{booktabs} 
\usepackage[normalem]{ulem} 
\begin{document}

\title{LLaPa: A Vision-Language Model Framework for Counterfactual-Aware Procedural Planning}

\author{Shibo Sun}
\affiliation{%
  \institution{Harbin Institute of Technology}
  \city{Harbin}
  \country{China}}
\email{23s103177@stu.hit.edu.cn}

\author{Xue Li*}
\affiliation{%
  \institution{Harbin Institute of Technology}
  \city{Harbin}
  \country{China}}
\email{lixuecs@hit.edu.cn}

\author{Donglin Di}
\affiliation{%
  \institution{Li Auto}
  \city{Beijing}
  \country{China}}
\email{donglin.ddl@gmail.com}

\author{Mingjie Wei}
\affiliation{%
  \institution{Harbin Institute of Technology}
  \city{Harbin}
  \country{China}
}
\email{mjwei@ir.hit.edu.cn}

\author{Lanshun Nie}
\affiliation{%
  \institution{Harbin Institute of Technology}
  \city{Harbin}
  \country{China}}
\email{nls@hit.edu.cn}

\author{Wei-Nan Zhang}
\affiliation{%
  \institution{Harbin Institute of Technology}
  \city{Harbin}
  \country{China}
}
\email{wnzhang@ir.hit.edu.cn}

\author{Dechen Zhan}
\affiliation{%
  \institution{Harbin Institute of Technology}
  \city{Harbin}
  \country{China}}
\email{dechenzhanhit@gmail.com}

\author{Yang Song}
\affiliation{%
  \institution{University of New South Wales}
  \city{Sydney}
  \state{New South Wales}
  \country{Australia}}
\email{yang.song1@unsw.edu.au}

\author{Lei Fan*}
\affiliation{%
  \institution{University of New South Wales}
  \city{Sydney}
  \state{New South Wales}
  \country{Australia}}
\email{lei.fan1@unsw.edu.au}
\thanks{*Corresponding Authors}
\renewcommand{\shortauthors}{Sun, et al.}
\renewcommand\footnotetextcopyrightpermission[1]{}
\settopmatter{printacmref=false} 


\begin{abstract}
  While large language models (LLMs) have advanced procedural planning for embodied AI systems through strong reasoning abilities, the integration of multimodal inputs and counterfactual reasoning remains underexplored. To tackle these challenges, we introduce \textbf{LLaPa}, a vision-language model framework designed for multimodal procedural planning. LLaPa generates executable action sequences from textual task descriptions and visual environmental images using vision-language models (VLMs). Furthermore, we enhance LLaPa with two auxiliary modules to improve procedural planning. The first module, the Task-Environment Reranker (TER), leverages task-oriented segmentation to create a task-sensitive feature space, aligning textual descriptions with visual environments and emphasizing critical regions for procedural execution. The second module, the Counterfactual Activities Retriever (CAR), identifies and emphasizes potential counterfactual conditions, enhancing the model's reasoning capability in counterfactual scenarios. Extensive experiments on ActPlan-1K and ALFRED benchmarks demonstrate that LLaPa generates higher-quality plans with superior LCS and correctness, outperforming advanced models. The code and models are available \url{https://github.com/sunshibo1234/LLaPa}.
\end{abstract}


\maketitle
\newcommand{\blue}[1]{\textcolor{blue}{#1}}
\section{Introduction}
\begin{figure}[t]
  \centering
  \includegraphics[width=1\linewidth]{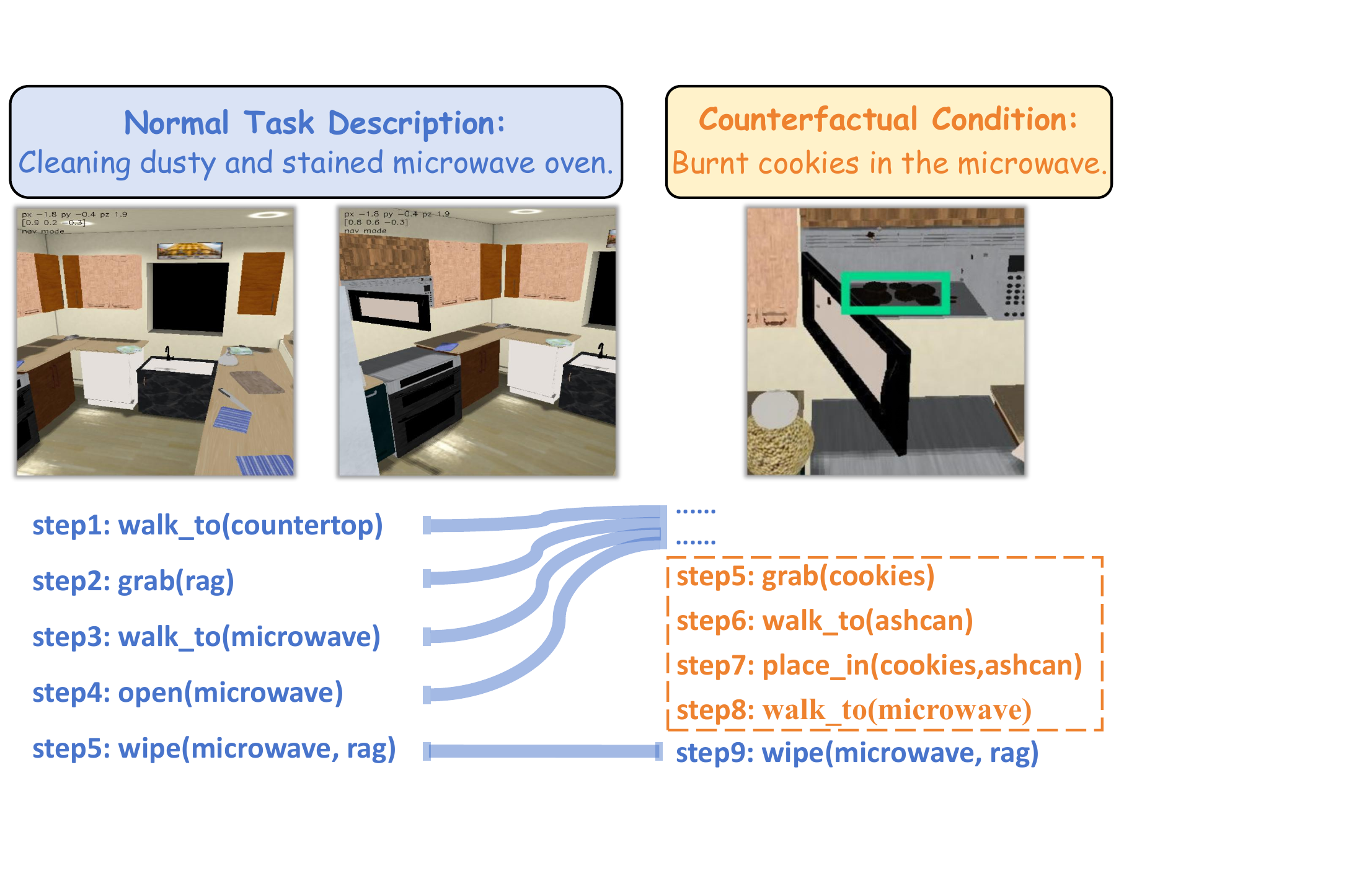}
  \caption{An example of procedural planning. Environmental factors and counterfactual activities (\textit{e.g.}, burnt cookies in the microwave oven) require additional reasoning and adaptation in the generated action plan.}
  \label{fig:intro}
\end{figure}

Recent advances in procedural planning have demonstrated significant progress in developing embodied agents capable of executing complex tasks by integrating both text instructions and visual information understanding \cite{francis2022core,li2023behavior,chang2020procedure-pp}. 
Procedural planning constitutes a structured reasoning process that decomposes high-level goals into coherent and logical action sequences, enabling an embodied system like a robot platform to accomplish real-world tasks. 
In this context, visual inputs help agents recognize object locations and possible accessibility, thereby defining the physical constraints on feasible actions.
These constraints often require the adaptation of dynamic plans to accommodate unexpected environmental states or deviations of tasks, which are collectively referred to as counterfactual conditions \cite{su2024actplan-dataset,holtman2021counterfactual}, as illustrated in Figure \ref{fig:intro}.
The dual demands of simultaneously possessing precise spatial reasoning capabilities and flexible adaptation mechanisms highlight the need for procedural planning systems.

Recently, various architectures based on pretrained foundation models \cite{huang2022inner-llm-intro,shah2023lm-intro,zitkovich2023rt-intro,firoozi2023foundation-intro,su2021prioritized} have been explored to enhance procedural planning. Existing approaches can broadly be categorized into two paradigms based on the predominant incorporated modality: Vision-Language Models (VLMs) and text-centric models.
Specifically, VLM-based approaches focus on fusing multimodal data (\textit{e.g.}, images or video frames) by integrating visual cues alongside textual descriptions to improve spatial reasoning and task adaptation.
For example, PaLM-E \cite{driess2023palm} is the first large-scale embodied multimodal model that directly extracts sensor-based features from real-world robot data \cite{zhang2024review}. EmbodiedGPT \cite{mu2023embodiedgpt-pp} employs an Embodied-Former that uses embodied queries and cross-attention between visual and textual features, selectively transferring information for more effective understanding and control in procedural planning.
In contrast, text-centric models, such as Plasma \cite{brahman2024plasma} and E2CL \cite{wang2024e2cl-pp3}, rely primarily on textual inputs. These models emphasize natural language instructions or dialogue, and often employ replanning or step-by-step refinement strategies to iteratively correct and refine action sequences throughout the procedural planning process.

However, while VLM-based methods are capable of processing multimodal inputs, their generic ability often tends to prioritize overall image semantics without establishing explicit interactions between textual tasks and visual environments during feature extraction. This can cause the model to overemphasize irrelevant objects that are not required for the current task. For example, in a microwave oven cleaning task, they may focus on the sink or jars instead of the microwave oven. Moreover, these models lack a clear mechanism for identifying and reasoning over counterfactual conditions. For example, when given an instruction such as ``\textit{if the microwave contains burnt cookies, discard them first,}'' these VLM-based models fail to associate the hypothetical condition (``\textit{if the microwave contains burnt cookies}'') with the relevant target objects (``\textit{burnt cookies}''), thereby degrading the reliability of the resulting action sequences.
Similarly, text-centric models have shown the ability to process counterfactual instructions by iteratively updating or refining plans through text-based reasoning. However, they lack visual grounding, which is essential for effective task execution in visually rich environments. As a result, while these models may correctly detect the need for replanning, their overall performance is hindered by the absence of visual context integration.

To overcome these challenges, we propose two key strategies.
First, to address the lack of interaction between tasks and objects, we introduce a method for images based on the objects explicitly mentioned in the task description. This establishes a tighter correspondence between textual instructions and environmental information. 
We then rerank the original image embeddings, enabling the model to prioritize the most task-relevant visual information during inference.
Second, to effectively handle counterfactual conditions, we detect such conditions in a single inference step, dynamically selecting and emphasizing the crucial visual cues related to these conditions. These cues are then transformed into visually grounded prompts, guiding the VLM to reason more accurately about counterfactual activities.

To this end, we propose LLaPa, a VLM-based framework for procedural planning, designing two specialized auxiliary modules, the Task-Environment Reranker (TER) and the Counterfactual Activities Retriever (CAR). 
Specifically, the TER module, inspired by task-oriented segmentation, leverages a semantic segmentation model to generate binary masks aligned with task-relevant objects. These masks are then incorporated into a self-attention mechanism to rerank visual features, ensuring that the model focuses on objects truly essential for task completion. 
Meanwhile, the CAR module employs a trained classifier to parse counterfactual conditions from task instructions, selects the corresponding visual tokens, and integrates these enhanced features into the VLM. This allows the model to reason over counterfactual conditions without requiring additional planning loops.
Compared to previous models \cite{brahman2024plasma,wang2024e2cl-pp3}, our framework achieves automatic identification and emphasis of counterfactual conditions in a single forward pass, significantly improving the ability and reliability to handle counterfactual activities.
By cascading these specialized modules with a VLM backbone, LLaPa can generate higher-quality action sequences that are better suited to complex scenarios. 
\begin{figure*}[t]
  \centering
  \includegraphics[width=1\linewidth]{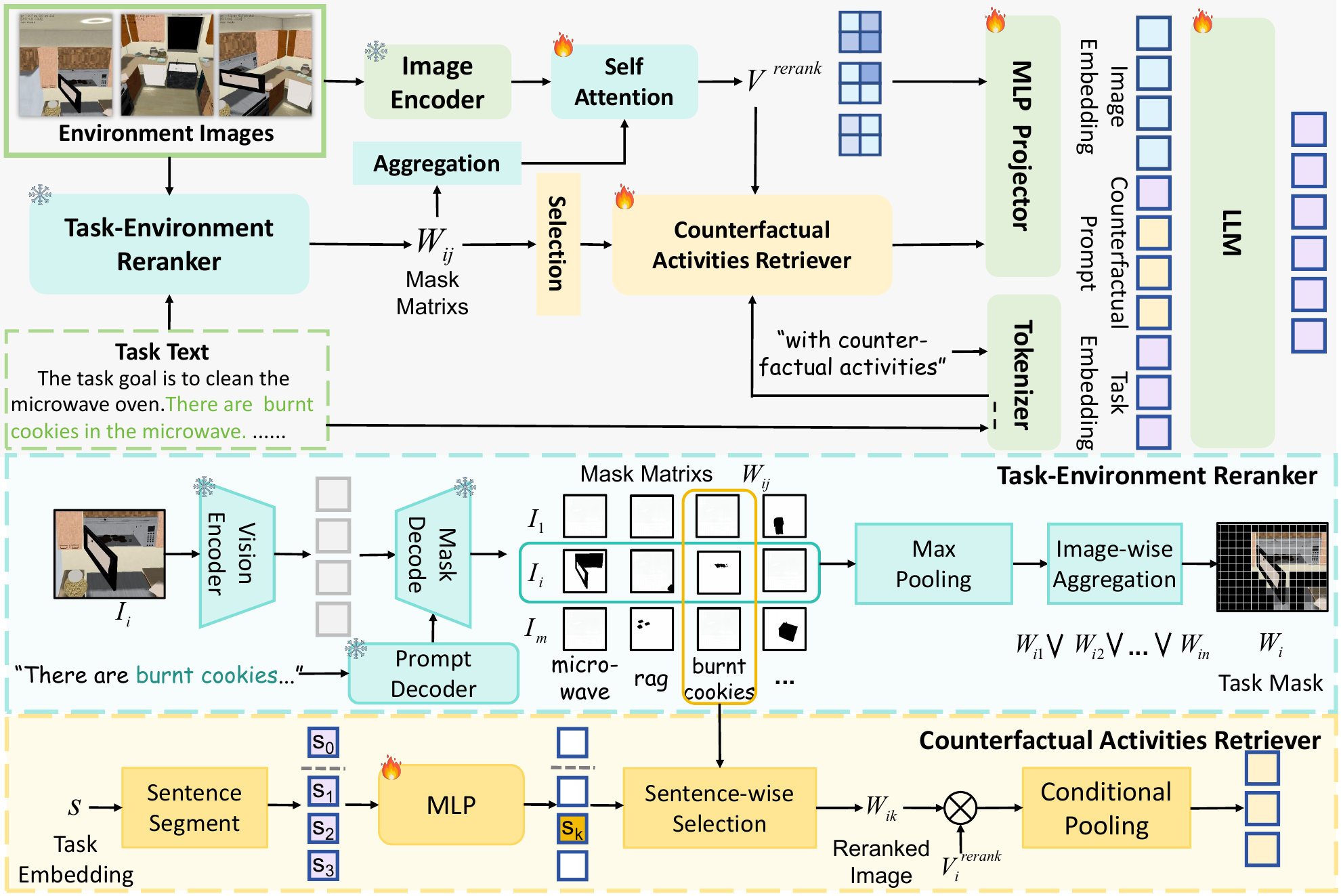}
  \caption{Overview of the LLaPa architecture. It consists of three components: the VLM backbone, the Task-Environment Reranker (TER), and the Counterfactual Activities Retriever (CAR). TER reranks the image embeddings to emphasize task-relevant objects, while CAR automatically detects counterfactual conditions and selects corresponding counterfactual visual tokens. Additionally, guided prompts are introduced with counterfactual visual tokens to further enhance the model’s capability in understanding and reasoning over counterfactual tasks.}
  \label{fig:overview}
\end{figure*}

Our key contributions are summarized as follows:
\begin{itemize}[topsep=0pt, parsep=0pt, itemsep=0pt]
\item We propose LLaPa, the first VLM-based framework that simultaneously addresses multimodal visual-text alignment and counterfactual adaptation for procedural planning.
\item We design the Task-Environment Reranker, which incorporates task-oriented segmentation principles to construct task-sensitive visual representations, effectively guiding the model toward task-critical elements. 
\item We develop a Counterfactual Activities Retriever, which leverages semantic constraint detection to automatically identify and emphasize counterfactual-related tokens.
\item Comprehensive experiments on the ActPlan-1K \cite{su2024actplan-dataset} and ALFRED \cite{shridhar2020alfred-benchmark} benchmarks demonstrate that LLaPa achieves state-of-the-art performance. Extensive qualitative analysis further reveals its superior robustness in handling complex scenarios involving counterfactual conditions. We show that transferring our proposed modules to various VLMs achieves significant gains, outperforming widely-used fine-tuning strategies \cite{hu2022lora}.
\end{itemize}

\section{Related Work}
\subsubsection*{\textbf{Vision-Language Models (VLMs).}}
Recent VLMs have shown impressive visual understanding capabilities \cite{li2024llava-VLM1,wang2024qwen2-VLM1,team2023gemini-VLM1,wu2024deepseek-VLM1,openai2023gpt4o-VLM1,liu2024llavanext-VLM1,cheng2024videollama2-VLM1,chen2024far-internvl2-VLM1,fan2025manta}. Among them, LLaVA \cite{liu2024improved-VLM4} first introduced the concept of \textit{visual instruction tuning}, sparking a series of subsequent developments \cite{zhang2024review}. Current research on general-purpose VLMs covers multiple fronts: i) Improving perceptual accuracy by processing high-resolution image inputs. Some models employed image slicing techniques to accommodate larger input sizes \cite{lin2023sphinx-VLM2,dai2024nvlm-VLM2}, while others incorporate specialized high-resolution encoders \cite{li2024mini-VLM2,luo2024feast-vlm2}; ii) Refining pre-training approaches and diversifying fine-tuning datasets \cite{lin2024vila-VLM3,xue2024xgen-VLM3}; and iii) Extending capabilities beyond single-image scenarios to areas~\cite{fan2025grainbrain,li2024llava-VLM4,liu2024improved-VLM4}.

Our work continues this line of research by extending VLM capabilities for procedural planning,  designing auxiliary modules that address real-world planning needs by aligning environment and instructions while handling counterfactual conditions.

\subsubsection*{\textbf{Procedural Planning.}}
The use of LLMs and VLMs in embodied agent tasks has garnered increasing attention, as they can formulate procedural plans endowed with actionable and commonsense knowledge. Text-centric methods typically rely on symbolic representations like PDDL to specify and reason about tasks \cite{silver2022pddl-pp1,guan2023leveraging-pp1,silver2024generalized-pp1,wu2024detecting}, or adopt chain-of-thought \cite{wei2022chain} and iterative correction strategies to build planning pipelines for more complex instructions \cite{lu2024strux-pp2,roy2024flap-pp2,zhou2024isr-pp2}. Some of these approaches support replanning or error correction when encountering counterfactual conditions \cite{wang2024e2cl-pp3,brahman2024plasma,yao2023react}. Meanwhile, VLM-based approaches integrate visual cues into the planning process, enabling agents to better interpret their surroundings and generate more contextually grounded plans \cite{ahn2024autort-pp4,lu2023multimodal-pp4,song2023llmplanner-pp4,mu2023embodiedgpt-pp}.

Previous work has not simultaneously addressed multimodal perception and counterfactual conditions, whereas our approach tackles these challenges by reranking visual features and selecting the tokens relevant to counterfactual conditions.

\subsubsection*{\textbf{Task-oriented Segmentation.}}
Image segmentation focuses on instance and semantic segmentation\cite{wang2021end-tos1,wang2023cut-tos1,cheng2021per-tos1,he2023fastinst-tos1,fan2023av4gainsp}. Recent research has moved beyond purely visual cues by incorporating textual queries, as seen in tasks such as referring expression segmentation or affordance detection \cite{lu202012-tes,yang2025learning,yang2024learning}. Building upon this, task-oriented segmentation (TOS) further advances the field by focusing on identifying objects relevant to a specific goal \cite{ravi2024sam2segmentimages,ren2024grounding,ren2024grounded-tos,zhu2025interpretable,yang2025dsdnet}. Compared to conventional segmentation, TOS not only demands accurate scene parsing but also requires goal-directed reasoning to identify objects essential for accomplishing a given task \cite{sawatzky2019object-tos3,fan2023identifying}.

We are the first to introduce TOS into a procedural planning framework, capitalizing on its ability to effectively link tasks and objects, thereby enabling the model to prioritize the most task-relevant visual information during inference.
\section{Method}

\subsection{Overview}
\subsubsection*{\textbf{Problem Definition.}}
We define the multimodal procedural planning problem as follows. Let 
\(\bm{S} = \{ S_{\text{goal}}, S_1, S_2, \dots, S_n \}\)
be a set of natural language sentences, where \(S_{\text{goal}}\) specifies the task goal and each \(S_i(i \in \{1,\cdots, n\})\) provides additional contextual information or conditions. Let 
\(\bm{I} = \{ I_1, I_2, \dots, I_m \}\)
be a set of environmental images, each of size \(H\!\times\!W\!\times\!3.\) The objective is to generate an ordered sequence of actions 
\(\bm{A} = \{ A_1, A_2, \dots, A_t \}\), 
where each action \(A_j(j \in \{1,\cdots, t\})\) is represented as a predicate-object combination like \(\mathrm{action}\) \((\mathrm{object})\).

To account for counterfactual factors that may alter the baseline plan, we decompose the action sequence as 
\(\bm{A} = [\bm{A}_{norm}, \bm{A}_{ctrf}]\).
Here, \(\bm{A}_{norm}\) represents the primary plan derived from the core goal, and \(\bm{A}_{ctrf}\) consists of additional or modified steps triggered by counterfactual conditions. Specifically, we further define a sub-function 
\( \alpha_{cls}(\bm{S}) = [\bm{S}_{norm}, \bm{S}_{ctrf}],\)
which identifies and separates the normal instructions \(\bm{S}_{norm}\) from the counterfactual information within the textual descriptions. Once detected, these counterfactual conditions can introduce new actions or modify existing actions in the final plan.
The final sequence of actions is generated as: \(\bm{A}= \phi\bigl([\bm{S}_{norm}, \bm{S}_{ctrf}],\, \bm{I}\bigr),\), where \(\phi\) is the mapping we aimed to learn,
indicating that both textual descriptions and environmental images are jointly integrated to produce the complete plan \(\bm{A}\).

\subsubsection*{\textbf{LLaPa Overview.}} The LLaPa model consists of three core components: the VLM backbone, the Task-Environment Reranker (TER), and the Counterfactual Activities Retriever (CAR).

The VLM backbone includes an image encoder, a text encoder, an MLP projector, and a large language model (LLM), as shown in Figure~\ref{fig:overview}. 
The backbone first processes the raw images \(\bm{I}\) and textual task descriptions \(\bm{S}\). The image encoder employs the Vision Transformer (ViT) architecture \cite{han2022survey-vit} with a patch size of \(P\), which segments each input image into \(P^2\) patches and maps them to \(C\)-dimensional vectors, resulting in image features \(v \in \mathbb{R}^{m \times P^2 \times C}\). 
Concurrently, the text encoder transforms the task description \(T\) into an embedding \(s \in \mathbb{R}^{L \times C}\), where \(L\) indicates the length of the text sequence. The outputs from both encoders are passed through the MLP projector, which aligns the multimodal features before feeding them into the LLM. The LLM produces a structured action sequence for coherent procedural planning.

In this process, TER refines the raw image embeddings \(v\in\mathbb{R}^{m\times P^2\times C}\) obtained from the VLM backbone by incorporating both the original images \(\bm{I}\) and textual descriptions \(\bm{S}\). It utilizes spatial weight maps \(W \in \{0,1\}^{m\times P^2\times 1}\), generated via promptable visual segmentation, which serves as attention masks in a self-attention mechanism. By highlighting task-relevant regions, the TER outputs refined embeddings \(v^{\text{rerank}}\in \mathbb{R}^{m\times P^2\times C}\) that guide the model toward more focused and goal-aligned procedural planning.

\begin{figure}[t]
  \centering
  \includegraphics[width=1\linewidth]{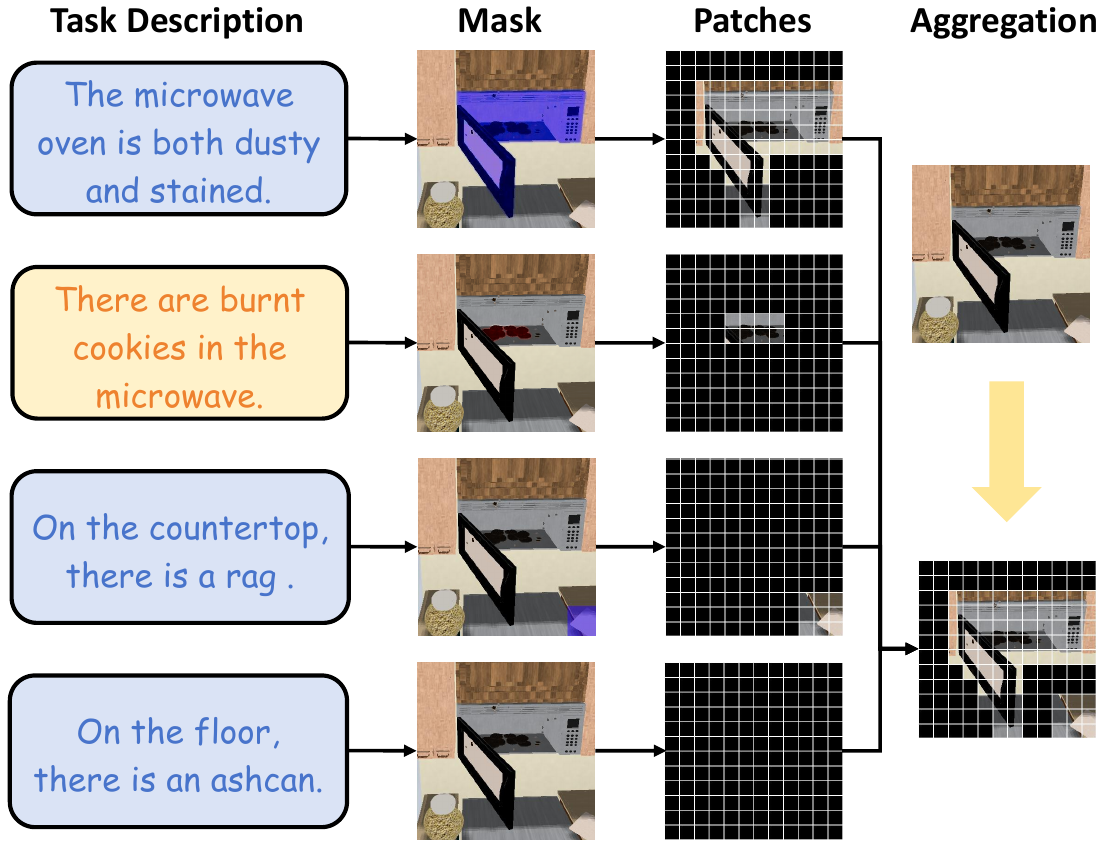}
  \caption{An example of task-driven segmentation. Although the jar in the lower-left corner of the image is a prominent visual element in the original image, it is excluded due to its irrelevance to the given task. Moreover, in the absence of an ashcan, the corresponding mask matrix contains only zeros.}
  \label{fig:reranker}
\end{figure}

CAR takes the textual embedding \(s\), the mask matrix \(W\), and the reranked image embeddings \(v^{\text{rerank}}\). By identifying counterfactual conditions within \(s\) and referencing the visual regions provided by \(W\), it extracts counterfactual-relevant visual tokens \(v^{\text{cf}}\) from \(v^{\text{rerank}}\). This process enables the model to attend to visual regions susceptible to unexpected or conditional scenarios, enhancing its ability to handle counterfactual reasoning.

Finally, the reranked visual tokens \(v^{\text{rerank}}\) and the counterfactual tokens \(v^{\text{cf}}\) are passed through the MLP projector and are then concatenated with the textual embedding, forming a unified multimodal representation that is fed into the LLM to produce an action sequence capable of addressing both normal and counterfactual conditions.

\subsection{Task-Environment Reranker}

While existing VLM-based approaches \cite{brahman2024plasma,mu2023embodiedgpt-pp} can associate textual instructions with environmental inputs, they often fail to align task-critical objects when counterfactual factors are involved, resulting in incomplete or inaccurate plans. 
Inspired by task-oriented segmentation \cite{ravi2024sam2segmentimages,ren2024grounded-tos}, which aims to segment task-relevant regions in complex scenes based on semantic descriptions, we propose a reranking mechanism to establish an explicit alignment between task content and visual environment. This allows the model to focus more on objects necessary for completing the task. 

Specifically, a frozen promptable segmentation model is employed, which consists of three components: a vision encoder that extracts multi-scale feature maps from the input images \(\bm{I}\); a prompt encoder that transforms textual descriptions \(\bm{S}\) into embeddings; and
a mask decoder that aligns the text and visual features to generate task-relevant segmentation masks \( M_{ij} \in \{0, 1\}^{H \times W \times 1} \). 
Given an image–description pair, since the VLM backbone operates on patch-level inputs, we apply an adaptive max pooling operation to convert \( M_{ij} \) into \( P^2 \) grids (where each grid has a size of \( \frac{H}{P} \times \frac{W}{P} \)), and the maximum pixel value within each grid is used to obtain weight matrix \( W_{ij} \in \{0, 1\}^{P \times P \times 1}\). 
These weight maps are then integrated into a self-attention layer, resulting in reranked visual features \(v^{\text{rerank}}\) that emphasize both standard and counterfactual task-relevant regions. As illustrated in Figure \ref{fig:reranker}, this process enables the model to focus only on the objects relevant to the task. By transforming pixel-level segmentation results into patch-level attention weights that are spatially aligned with the ViT feature map, it preserves the geometric structure of task-critical regions.

For multi-image inputs, we construct global spatial weights through image-level aggregation. Given an environmental input containing \(m\) images, the patch-level weights \(\{W_{i1}, W_{i2}, \dots, W_{in}\}\) corresponding to each textual constraint are combined via a bitwise OR operation to form a comprehensive weight matrix \(W_i\) for each, defined as:
\begin{equation}
    W_i = \bigvee_{j=1}^n W_{ij} ,
\end{equation}
where \(\bigvee\) denotes the bitwise OR operation. The weights of all images are concatenated to form the global attention mask matrix \( W \). This mask is integrated into the self-attention mechanism \cite{vaswani2017attention} of the visual encoder as follows:
\begin{equation}
\mathbf{Attn}(v, W) = \text{softmax}\left(\frac{QK^\top}{\sqrt{d}} + \log(W)\right)V,
\end{equation}
where \( Q = W_Q v \), \( K = W_K v \) and \( V = W_V v \) are the queries, keys, and values; \( W_Q \), \( W_K \) and \( W_V \) are learnable projection matrices; and \( d \) is the dimensionality of the keys. 
Through this weighted self-attention mechanism, the reranked features \(v^{\text{rerank}}\) are obtained by reweighting image patches based on task relevance. This process not only retains the hierarchical feature extraction of the ViT backbone but also ensures that the model attends to task-critical objects, especially under counterfactual conditions. The resulting spatial weight mask \(W\) and the refined features \(v^{\text{rerank}}\) serve as informative inputs for the counterfactual activities retriever to more effectively detect and handle unexpected changes or alternative actions within the task context.

\subsection{Counterfactual Activities Retriever}

This module identifies and highlights textual clauses that introduce potential counterfactual conditions, and then projects them onto the visual domain to extract relevant tokens. 
It takes three inputs: the textual embedding \(s\), the reranked visual features \(v^{\mathrm{rerank}}\), and the spatial weight masks \(W\). 
Internally, it comprises four sub-components: a sentence segmenter that partitions the input \(s\) into individual clauses, a classifier that determines relevant spatial masks from \(W\) based on the classification results, and a conditional pooling stage that combines these selected masks with \(v^{\mathrm{rerank}}\) to extract relevant features \(v^{\mathrm{cf}}\), which are set of tokens capturing the visual regions and textual cues critical for counterfactual reasoning.

To detect potential counterfactual conditions, the textual embedding is segmented into \(n+1\) clauses \(\{s_0, s_1, \dots, s_n\}\) based on sentence delimiters. Here, each \(s_i\) is the embedded representation of the corresponding textual clause \(S_i\), and \(s_0\) represents the embedding of the main goal \(S_{\mathrm{goal}}\). We define a classifier function \(\alpha_{cls}\), implemented by an MLP followed by a \(\text{sigmoid}\) activation:
\begin{equation}
p_k = \frac{1}{1 + \exp\left( -\alpha_{cls}([s_0; s_k]) \right)},
\label{eq:important}
\end{equation}
where \(p_k \in [0,1]\) indicates the likelihood that the \(k\)-th clause corresponds to a counterfactual constraint.
The set of indices of identified counterfactual activities is denoted as  \( \mathcal{K}_{\text{cf}} \). This step enables the model to dynamically identify textual conditions that may alter the execution path.

To enhance visual features based on detected counterfactual activities, for each \( k \in \mathcal{K}_{\text{cf}} \), the corresponding patch-level weights \( W_{ij} \) are first aggregated into a per-image binary counterfactual weight matrix \( W^{\text{cf}}_i \in \{0,1\}^{P^2 \times 1} \). This matrix explicitly marks task-critical regions influenced by counterfactual conditions within the \( i \)-th image. The reranked visual features \( V^{\text{rerank}}_i \) are then element-wise multiplied with \( W^{\text{cf}}_i \) to produce masked features \( v'_i = v^{\text{rerank}}_i \otimes W^{\text{cf}}_i \), where only the regions aligned with the counterfactual conditions are preserved. To extract task-critical features while suppressing environmental noise, each \( P \times P \) visual block is further partitioned into \( K \times K \) sub-grids. A conditional pooling operation is then applied to each sub-grid \( g \in \{1, \dots, K^2\} \):  
\begin{equation}
V^{\text{cf}}_{i,g} = 
\begin{cases} 
\frac{1}{|\Omega_g|} \sum_{p \in \Omega_g} v'^{(p)}_{i} & \text{if } \Omega_g \neq \emptyset, \\
\mathbf{0} & \text{otherwise.}
\end{cases},
\end{equation} 
where \( \Omega_g = \mathcal{G}_g \cap \mathcal{N}(W^{\text{cf}}_i = 1) \), \( \mathcal{G}_g \) denotes the indices of the \( g \)-th sub-grid, and \( \mathcal{N}(W^{\text{cf}}_i = 1) \) represents the non-zero positions in \( W^{\text{cf}}_i \). Finally, all sub-grid embeddings \(v^{\text{cf}}_{i,g}\) are concatenated to obtain the per-image counterfactual feature set \(v^{\text{cf}}_i\), which compactly represents the visual tokens relevant to the identified counterfactual conditions.

Compared to simple average pooling \cite{mu2023embodiedgpt-pp}, conditional pooling aggregates information only from regions related to counterfactual activities, thereby avoiding interference from irrelevant areas, and more effectively preserving task-critical features. This operation generates compact counterfactual features \( v^{\text{cf}}_i \in \mathbb{R}^{K^2 \times C} \), achieving a transformation from fine-grained visual responses to semantic representations that are specifically aligned with counterfactual conditions.
\newcommand{\best}[1]{\textbf{#1}}
\newcommand{\second}[1]{\uline{#1}}

\begin{table*}[!t]
\caption{Quantitative comparison of the models on ActPlan-1K (including both counterfactual activities and normal activities), ALFRED, and MFE-ETP. The best and the second-best results are highlighted in bold and underlined, respectively. The size of DeepSeek-VL2 refers to its activated parameters. The ``8B'' listed for LLaPa refers to the InternVL2 backbone, as the additional parameters introduced by TER and CAR are negligible (far less than 0.2 billion).}
\centering
\resizebox{\textwidth}{!}{
\begin{tabular}{lcc|ccc@{\hspace{8pt}}|ccccccc}
\toprule
\multirow{2}{*}{Model} & 
\multirow{2}{*}{Type} &
\multirow{2}{*}{Size} &
\multicolumn{3}{c|}{ActPlan-1K (ctrf.) \cite{su2024actplan-dataset}} &
\multicolumn{3}{c}{ActPlan-1K (norm.) \cite{su2024actplan-dataset}} &
\multicolumn{3}{c}{ALFRED \cite{shridhar2020alfred-benchmark}} &
\multicolumn{1}{c}{MFE \cite{zhang2024mfe-benchmark}} \\
\cmidrule(lr){4-6} \cmidrule(lr){7-9} \cmidrule(lr){10-12} \cmidrule(lr){13-13}
& & & Exec.$\uparrow$ & LCS$\uparrow$ & Corr.$\uparrow$ & Exec.$\uparrow$ & LCS$\uparrow$ & Corr.$\uparrow$ & Exec.$\uparrow$ & LCS$\uparrow$ & Corr.$\uparrow$ & Acc.$\uparrow$ \\
\midrule
GPT-4o \cite{openai2023gpt4o-VLM1} & \multirow{2}{*}{Close-set} & - 
& \second{49.2} & 0.48 & 21.4
& 56.5 & 0.59 & \second{39.8}
& \best{90.4} & 0.60 & \second{47.4}
& \best{78.0} \\

Gemini-Pro-1.5 \cite{team2023gemini-VLM1} & & -
& 46.0 & 0.51 & 25.8
& 51.0 & 0.55 & 38.6
& 84.0 & \second{0.60} & 46.6
& \second{73.3} \\
\midrule
LLaVa-OV \cite{li2024llava-VLM1} & \multirow{5}{*}{Open-set} & 7B
& 37.3 & 0.49 & 24.6
& 50.6 & 0.55 & 35.5
& 78.2 & 0.42 & 29.8
& 30.5 \\

VideoLLaMA2 \cite{cheng2024videollama2-VLM1} & & 7B
& 30.2 & 0.43 & 20.3
& 40.6 & 0.48 & 31.9
& 73.1 & 0.57 & 43.9
& 46.3 \\

DeepSeek-VL2 \cite{wu2024deepseek-VLM1} & & 4.5B
& 34.9 & 0.44 & 23.8
& 38.2 & 0.50 & 24.4
& 72.1 & 0.54 & 32.1
& 31.8 \\

Qwen2-VL \cite{wang2024qwen2-VLM1} & & 7B
& 43.7 & 0.53 & 27.9
& \second{57.8} & \second{0.59} & 37.9
& 81.6 & 0.56 & 43.8
& 56.2 \\

InternVL2 \cite{chen2024far-internvl2-VLM1} & & 8B
& 44.9 & 0.52 & 26.1
& 52.8 & 0.57 & 38.2
& 81.9 & 0.58 & 45.3
& 49.7 \\

InternVL2(Plasma) \cite{brahman2024plasma} & & 8B
& 48.2 & \second{0.53} & \second{30.2}
& 53.9 & 0.58 & 39.0
& 81.8 & 0.58 & 45.7
& 50.1 \\
\midrule
Embodied-GPT \cite{mu2023embodiedgpt-pp} & \multirow{2}{*}{Specialized} & 7B
& 39.8 & 0.48 & 21.5
& 54.3 & 0.57 & 36.6
& 70.7 & 0.56 & 41.6
& 51.5 \\

LLaPa (Ours) & & 8B
& \best{53.2} & \best{0.57} & \best{36.1}
& \best{62.9} & \best{0.62} & \best{45.2}
& \second{85.3} & \best{0.62} & \best{48.6}
& 58.6 \\
\bottomrule
\end{tabular}
}
\label{tab:all}
\end{table*}

\subsection{Training and Inference}

\textbf{Training.} 
We adopt a two-stage training procedure. In the first stage, we focus on learning the counterfactual classifier. Only the classifier’s parameters are updated, while the rest of the network remains frozen. This allows the model to reliably detect textual clauses that introduce counterfactual conditions. In the second stage, we fine-tune the action-sequence generation pipeline, which includes the reranker’s self-attention layers, the MLP projector, and the LLM. This stage enables the overall system to produce coherent action sequences.

The overall training loss \(\mathcal{L}\) comprises two main parts: counterfactual classification loss \(\mathcal{L}_{\mathrm{cf}}\), which ensures accurate identification of clauses and action-sequence generation loss \(\mathcal{L}_{\text{LLM}}\), which optimizes the quality of the
generated action sequence. Specifically, we use a binary cross-entropy loss to measure the deviation between the classifier’s output \(p_k\) and the human-annotated label \(y_k\):

\begin{equation}
\mathcal{L}_{\mathrm{cf}} 
= -\frac{1}{N} \sum_{k=1}^{N}
\Bigl[
   y_k\,\log(p_k)
   +\bigl(1 - y_k\bigr)\,\log\bigl(1 - p_k\bigr)
\Bigr].
\label{eq:cf_loss}
\end{equation}

\noindent The action-sequence generation loss adopts a standard language modeling loss, forcing the model to maximize the likelihood of the ground truth action sequence given the multimodal context:
\begin{equation}
\mathcal{L}_{\text{LLM}} = -\frac{1}{T} \sum_{t=1}^{T} \log P(a_t \mid s^{\text{input}}, a_{<t}),
\end{equation} 
where \( P(a_t \mid s^{\text{input}}, a_{<t}) \) is the probability of the correct token \( a_t \) given the input embeddings \( s^{\text{input}} \) and the previously generated tokens \( a_{<t} \), and \( T \) is the length of the target action sequence.



\textbf{Inference.} During the inference phase, counterfactual cues are explicitly injected into the decoding process. The reranked features \( v^{\text{rerank}} \) and the counterfactual tokens \( v^{\text{cf}} \) are aligned to the language embedding space through a shared MLP projection, and then concatenated with the text encoding \( s \) and a special prompt token \( s^{\text{cf}} \) before being input into the LLM:
\begin{equation}
s^{\text{input}} = \left[\text{Proj}(v^{\text{rerank}}); s^{\text{cf}}; \text{Proj}(v^{\text{cf}}); s\right].
\end{equation}
Here, \( s^{\text{cf}} \) represents the prompt embedding, such as ``\textit{With counterfactual conditions’ environmental features...}'', guiding the LLM to establish explicit associations between counterfactual conditions and action planning. Finally, the input is passed to the LLM for decoding, resulting in the action sequence \( \bm{A}\).

\section{Experiment}

\subsection{Setup}
\textbf{Datasets.}  
We evaluated our model on ActPlan-1K \cite{su2024actplan-dataset}, currently the only multimodal procedural planning benchmark that is known to support counterfactual activities. By introducing unconventional conditions, the dataset tests the adaptive planning capabilities of models. To further assess the zero-shot generalization, we additionally conducted evaluations on the ALFRED \cite{shridhar2020alfred-benchmark}, which is widely used for embodied planning tasks. To evaluate general task planning capabilities, we also included evaluations on the MFE-ETP \cite{zhang2024mfe-benchmark}. Detailed dataset information is provided in \textit{\textbf{Appendix}}.

\textbf{Implementation Details.}  
Our LLaPa model was based on the InternVL2-8B \cite{chen2024far-internvl2-VLM1} framework, where the visual encoder adopted the InternViT architecture with a patch size of \( P = 16 \), and the language model was based on InternVL2. For the Task-Environment Reranker module, we employed the Grounded-Segment Anything model \cite{liu2023grounding-tos,ren2024grounded-tos}. The self-attention layer follows the standard Transformer \cite{vaswani2017attention,su2022vitas}, with an 8-head attention mechanism applied for spatial feature reweighting. The counterfactual activities retriever sets the sub-grid division parameter \( K = 4 \), resulting in 16 counterfactual-related visual tokens extracted per image. 

The model training was divided into two independent stages. In the first stage, we used the entire COPLAN \cite{brahman2024plasma} dataset and the ActPlan-1K training set to train the components of the counterfactual activities retriever. This stage focuses on acquiring the ability to identify counterfactual activity clauses, with all other parameters frozen. In the second stage, we utilized the EgoCOT \cite{mu2023embodiedgpt-pp} dataset together with the ActPlan-1K training set. 
More training details are provided in \textit{\textbf{Appendix}}.

\textbf{Metrics.}  
Following previous work \cite{brahman2024plasma,song2023llmplanner-pp4,huang2022language-metrics,puig2018virtualhome-metrics}, we evaluated the quality of the generated action sequences from three perspectives: Executability, LCS, and Correctness.  
i) \textbf{Executability} measures whether the generated operations can be executed in the environment, reported as a percentage.
ii) \textbf{Longest Common Subsequence (LCS) } measures the action-level similarity between a
generated plan and a human-annotated one.
iii) \textbf{Correctness} assesses whether the generated actions accomplish the task expressed as a percentage. For the MFE-ETP question-answering dataset, we reported \textbf{Accuracy} as the evaluation metric.


\begin{figure*}[t] 
  \centering
  \begin{subfigure}[b]{0.48\textwidth} 
    \includegraphics[width=\linewidth]{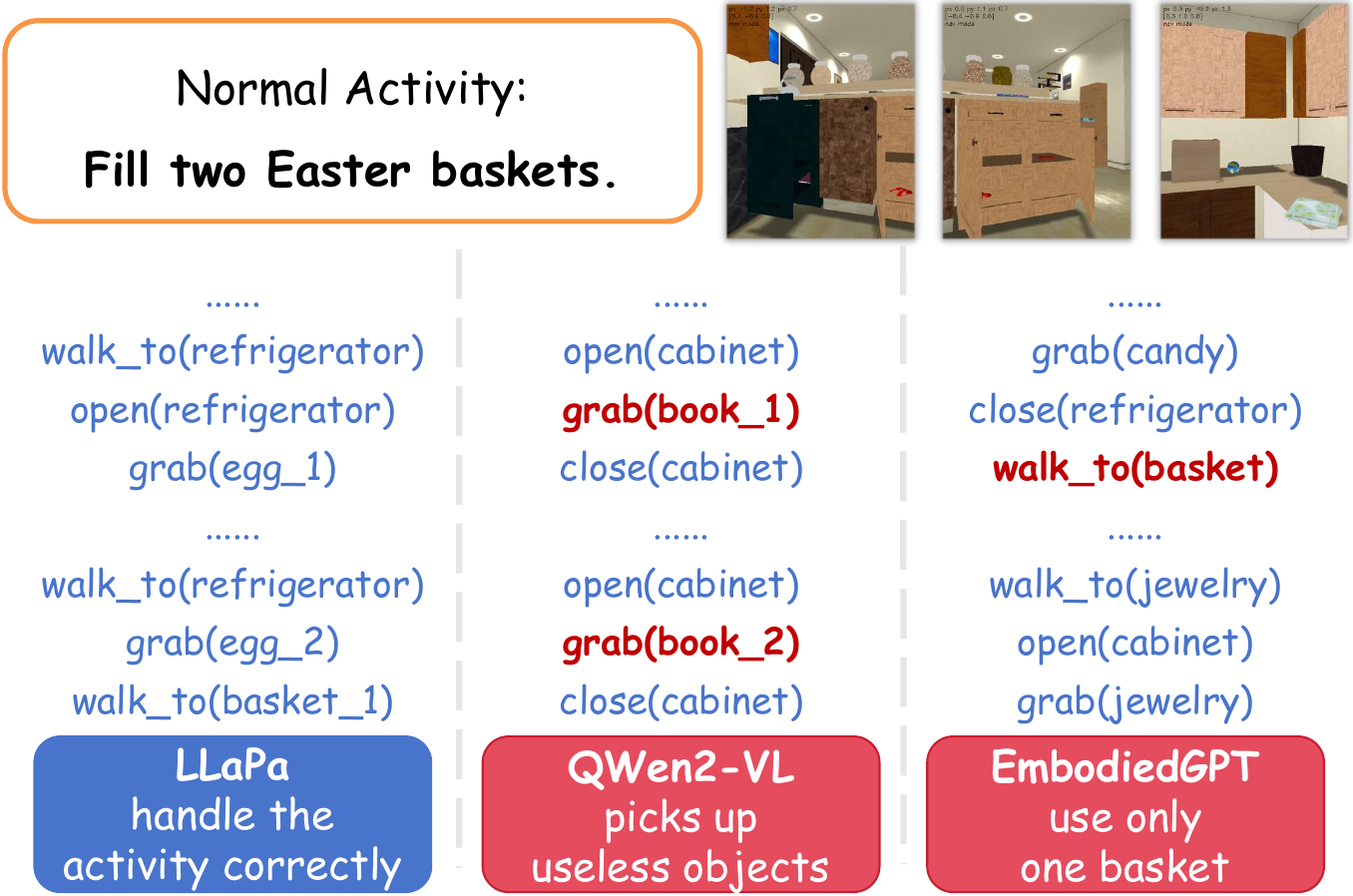}
    \caption{An example of a normal activity. LLaPa correctly processes activities, while other models generate tool hallucinations or incorrect environmental understanding.}
    \label{fig:case1}
  \end{subfigure}
  \hfill 
  \begin{subfigure}[b]{0.48\textwidth}
    \includegraphics[width=\linewidth]{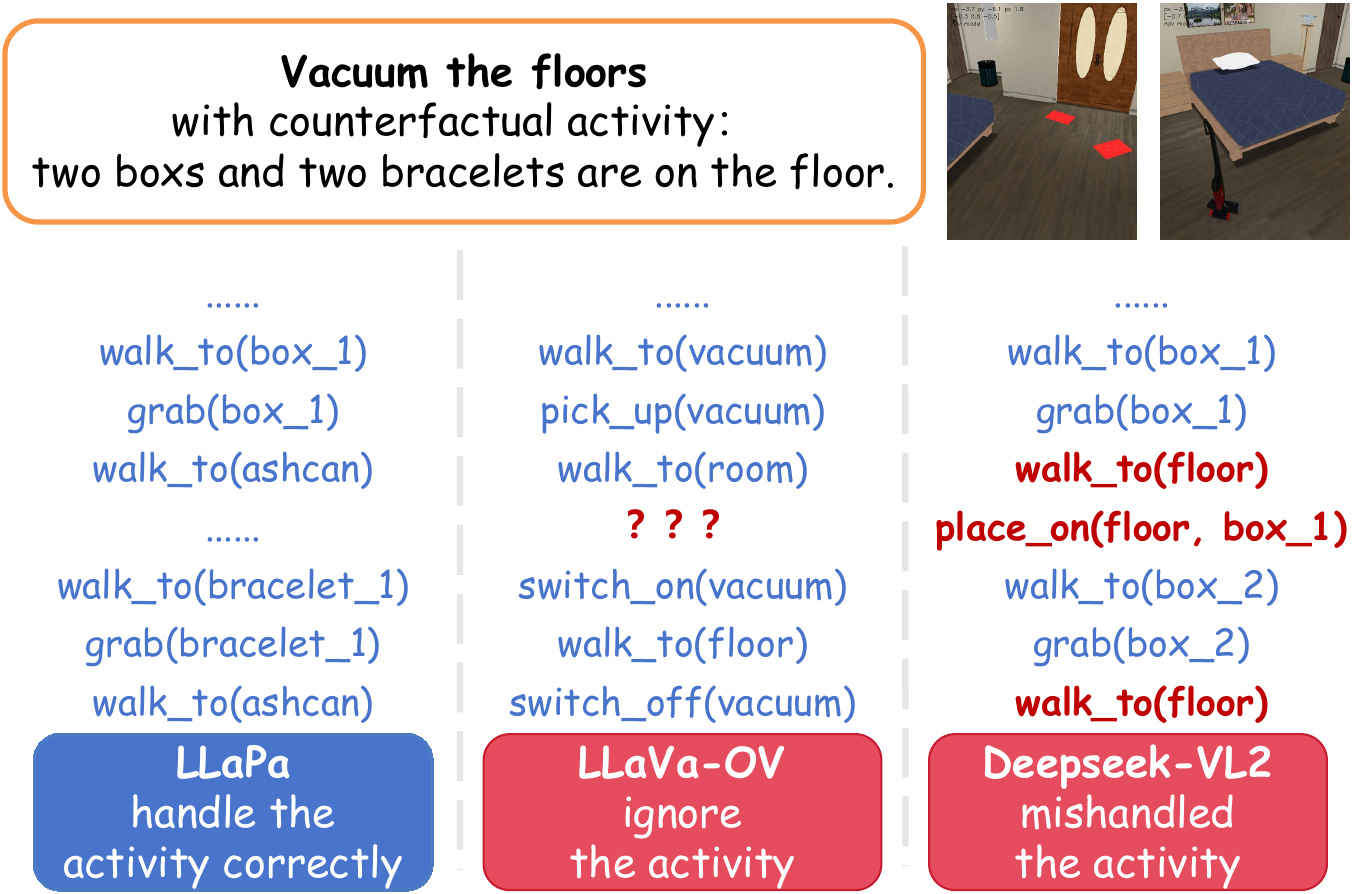}
    \caption{An example of a counterfactual activity. LLaPa correctly identifies and processes counterfactual conditions, while other models either fail to handle them properly or completely disregard such conditions.}
    \label{fig:case2}
  \end{subfigure}
  \caption{Qualitative comparison between LLaPa and other models.}
  \label{fig:cases}
\end{figure*}
\begin{figure}[t]
  \centering
  \begin{subfigure}[t]{\linewidth}
    \centering
    \includegraphics[width=\linewidth]{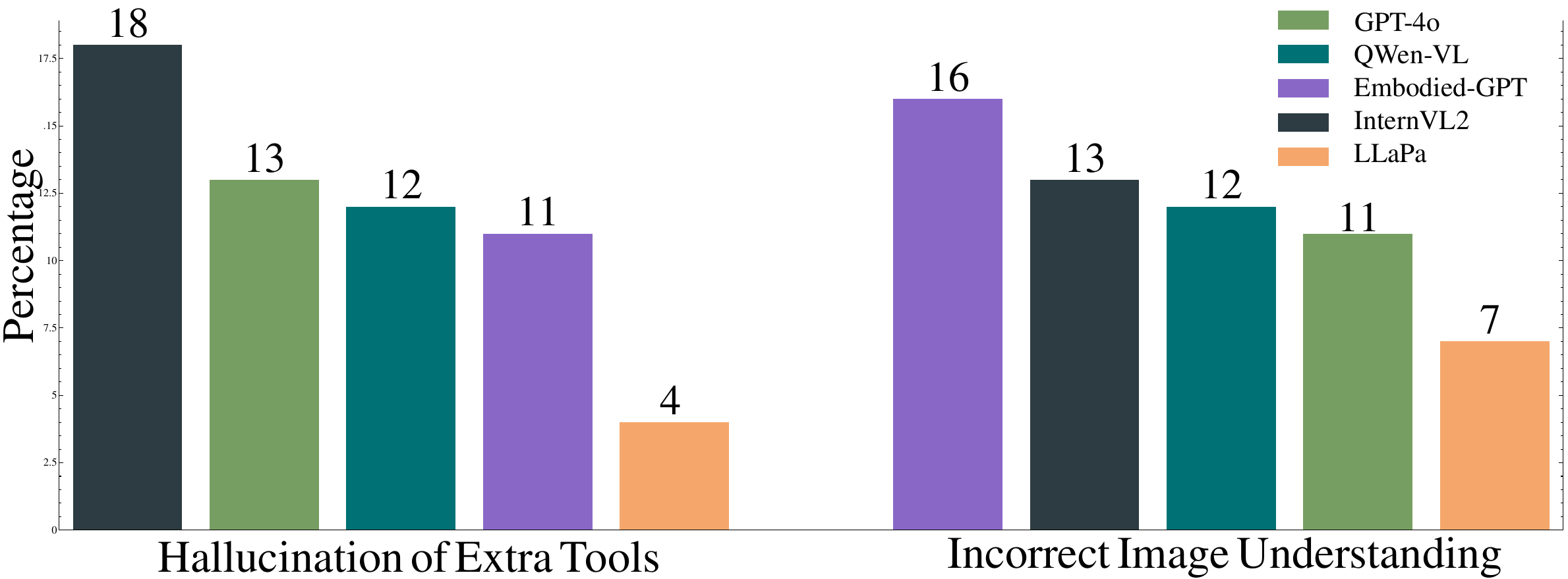}
    \caption{Results in ActPlan-1K(norm.).}
    \label{fig:model_performance_norm}
  \end{subfigure}
  \vspace{0.5cm} 
  \begin{subfigure}[t]{\linewidth}
    \centering
    \includegraphics[width=\linewidth]{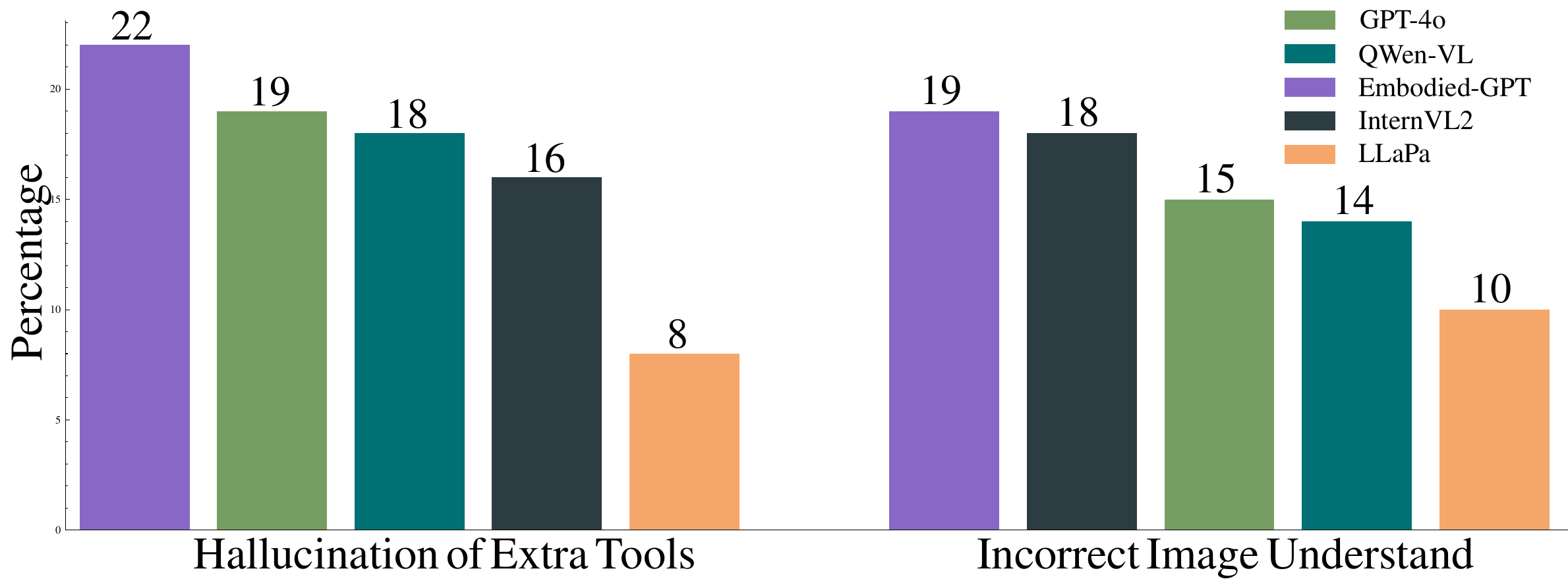}
    \caption{Results in ActPlan-1K(ctrf.).}
    \label{fig:model_performance_ctr}
  \end{subfigure}
  \caption{Error analysis. Compared to other models, LLaPa generates action sequences with significantly fewer errors in environmental understanding.}
  \label{fig:model_performance}
\end{figure}
\subsection{Comparisons}
Our experiments aimed to thoroughly evaluate the effectiveness of LLaPa under both normal and counterfactual procedural planning scenarios. To this end, we compared LLaPa with representative baselines across three datasets, conducting extensive qualitative and quantitative evaluations to assess each model’s ability in both normal and counterfactual activity planning.

\textbf{Baselines.}  
We selected three categories of advanced methods: (1) close-set VLMs, including GPT-4o \cite{openai2023gpt4o-VLM1} and Gemini-Pro-1.5 \cite{team2023gemini-VLM1}; (2) open-set VLMs, including LLaVA-OV \cite{li2024llava-VLM1}, VideoLLaMA2 \cite{cheng2024videollama2-VLM1}, DeepSeek-VL2 \cite{wu2024deepseek-VLM1}, Qwen2-VL \cite{wang2024qwen2-VLM1}, InternVL2 \cite{chen2024far-internvl2-VLM1}, as well as a custom baseline using InternVL with Plasma's\cite{brahman2024plasma} CoT-style reasoning; (3) specialized procedural planning model: Embodied-GPT \cite{mu2023embodiedgpt-pp}. All baseline models were reproduced under the same hardware conditions, with a maximum generation length limited to 512 tokens. To ensure fairness in the evaluation of ActPlan-1K \cite{su2024actplan-dataset}, we fine-tuned the open-set models using the ActPlan-1K training set.

\textbf{Quantitative Results.}  
As presented in Table \ref{tab:all}, for normal tasks (\textit{i.e.}, without counterfactual conditions), LLaPa demonstrated a clear advantage in all benchmarks. Specifically, it achieved 62.9\% Executability, 0.62 LCS, and 45.2\% Correctness on ActPlan-1K \cite{su2024actplan-dataset}, substantially outperforming other open-set and specialized models. 
We attribute these improvements to TER, which effectively filters out task-irrelevant content and emphasizes critical objects, thus reducing common misalignment issues found in generic vision language solutions.

In the counterfactual conditions, the result showed that LLaPa continued to demonstrate superior performance, achieving 53.2\% Executability, 0.57 LCS, and 36.1\% Correctness on ActPlan-1K \cite{su2024actplan-dataset}. In comparison, general-purpose VLMs such as GPT-4o \cite{openai2023gpt4o-VLM1}  and Qwen2-VL \cite{wang2024qwen2-VLM1} exhibited significantly lower Correctness (21.4\% and 27.9\%, respectively). This outcome confirmed the effectiveness of the CAR, which detects and highlights key variations in the task description and focuses on the most pertinent visual tokens. This targeted attention allows LLaPa to dynamically adjust its action sequence to unforeseen conditions.

\textbf{Qualitative Results.}  
We illustrated a comparison between LLaPa and other models, as shown in Figure \ref{fig:case1} and Figure \ref{fig:case2}. The qualitative analysis of the examples revealed that the LLaPa model consistently produced more coherent and contextually appropriate plans compared to other models. In particular, the model exhibits a superior ability to reason over dynamic task constraints, including the presence of counterfactual conditions.
These results highlight LLaPa's effectiveness and reliability in complex planning environments.

\begin{table*}[t]
    \centering
    \caption{\textbf{Left:} Ablation study on the numbers of counterfactual tokens. \textbf{Right:} Ablation study on the proposed components.}
    \label{tab:ablation_studies}
    \small
    
    \begin{minipage}[t]{0.30\linewidth}
    \centering
    \setlength{\tabcolsep}{8pt}
    \begin{tabular}{lccc}
    \toprule
    \multirow{2}{*}K & \multicolumn{3}{c}{ActPlan-1K (ctrf.)} \\
    \cmidrule(lr){2-4}
     & Exec. & LCS & Corr. \\
    \midrule
    K=2  & 49.2 & 0.51 & 32.5 \\ 
    K=4  & \best{52.8} & \best{0.57} & \best{36.4} \\  
    K=8  & 41.0 & 0.49 & 26.8 \\ 
    K=16 & 33.2 & 0.38 & 21.5 \\  
    \bottomrule
    \end{tabular}
    \end{minipage}
    \hfill
    \begin{minipage}[t]{0.68\linewidth}
    \centering
    \setlength{\tabcolsep}{8pt}
    \renewcommand{\arraystretch}{1.0}
    \begin{tabular}{lccccccccc}
    \toprule
     & \multicolumn{3}{c}{ActPlan-1K (ctrf.)} 
     & \multicolumn{3}{c}{ActPlan-1K (norm.)} 
     & \multicolumn{3}{c}{ActPlan-1K (total)} \\ 
    \cmidrule(lr){2-4} \cmidrule(lr){5-7} \cmidrule(lr){8-10}
     & Exec. & LCS & Corr.
     & Exec. & LCS & Corr.
     & Exec. & LCS & Corr. \\
    \midrule
    Full     
    & \best{53.9} & \best{0.57} & \best{36.1}
    & 62.7 & \best{0.62} & 45.2     
    & \best{59.4} & \best{0.60} & \best{42.4}   \\
    
    w/o CAR 
    & 47.6 & 0.51 & 30.7     
    & \best{63.8} & 0.62 & \best{45.8}
    & 58.3 & 0.58 & 40.8   \\
    
    w/o TER 
    & 44.2 & 0.49 & 28.4     
    & 52.4 & 0.57 & 39.1     
    & 49.6 & 0.54 & 35.5   \\
    
    Only SFT 
    & 36.5 & 0.45 & 20.7     
    & 50.3 & 0.52 & 37.3     
    & 46.7 & 0.49 & 31.6   \\
    \bottomrule
    \end{tabular}
    \end{minipage}
\end{table*}

\subsection{Impact of Modules on Error Reduction}

To evaluate the effectiveness of our newly integrated modules in reducing specific types of errors, we focused on errors related to the environment, which are crucial for accurate task execution.

We followed the methodology of ActPlan-1K\cite{su2024actplan-dataset} and conducted an error analysis by sampling 40 activity types. Our evaluation specifically targets two error categories: (1) \textbf{Hallucination of Extra Tools:} Errors where the model generates actions involving tools not present in the images or task descriptions.
(2) \textbf{Incorrect Image Understanding:} Errors arising from misinterpreting visual content, such as misjudging distances or the number of objects. 

Our approach significantly reduced targeted errors compared to other models, as shown in Figure \ref{fig:model_performance}. Our model achieved the lowest error percentage, particularly for counterfactual activities, with reductions of 8\% and 4\% respectively. The TER improved spatial alignment, reducing hallucination of extra tools, while the CAR enhanced contextual understanding by emphasizing counterfactual conditions.
\subsection{Ablation Study}

\textbf{Effect of Numbers of Counterfactual Tokens.}  
The ablation study on the parameter \( K \) (Table \ref{tab:reranker_strategies}) showed that the model achieved the best performance when \( K = 4 \), with 4.0\% higher executability and 3.6\% greater correctness compared to \( K = 2 \). 
We hypothesize that when \( K \) is too small, the granularity of feature extraction is insufficient, leading to a decline in the correctness of counterfactual tasks. 
Conversely, when \( K \) is too large, overly fine divisions introduce environmental noise, reducing structural consistency.

\textbf{Effect of Proposed Components.}  
We further investigated the effectiveness of TER and CAR, as shown in Table \ref{tab:reranker_strategies}. Removing the CAR module significantly resulted in a significant 5.4\% decrease in the correctness of counterfactual tasks, confirming the necessity of explicit constraint modeling. Removing TER's segmentation strategy decreased object state sensitivity and caused a 9.8\% executability decline. Without both modules, the model failed in counterfactual tasks, demonstrating the irreplaceability of multimodal feature decoupling. TER alone showed limited adaptability to dynamic conditions. Our LLaPa model, integrating both TER and CAR, achieved optimal performance.

\textbf{Effect of TER Strategies.}
We also compared different strategies for the TER module on ActPlan-1K. In addition to the default bitwise OR operation, we examined:
(1) summing and normalizing all mask matrices;
(2) computing the similarity between textual and image-patch embeddings;
(3) applying no aggregation strategy.

\begin{table}[t]
    \centering
    \caption{Comparison of different TER strategies on ActPlan-1K (Total) and ALFRED. 
    “Sim.” denotes text-image similarity, and “/” denotes no strategy applied.}
    \label{tab:reranker_strategies}
    \begin{tabular}{cc|cccccc}
    \toprule
    \multicolumn{2}{c|}{Strategies} 
    & \multicolumn{3}{c}{ActPlan-1K (total)} 
    & \multicolumn{3}{c}{ALFRED} \\
    \cmidrule(lr){1-2}\cmidrule(lr){3-5}\cmidrule(lr){6-8}
    Mask & Aggr. 
    & Exec. & LCS & Corr.
    & Exec. & LCS & Corr. \\
    \midrule
    TOS & OR 
        & 58.9 & \textbf{0.60} & \textbf{42.0} 
        & \textbf{85.2} & 0.62 & \textbf{48.9} \\
    TOS & Sum.
        & 54.9 & 0.56 & 37.1
        & 83.4 & \textbf{0.63} & 45.1 \\
    Sim. & Sum.
        & \textbf{59.2} & 0.59 & 39.4
        & 78.2 & 0.57 & 44.3 \\
    / & /
        & 47.1 & 0.50 & 32.4
        & 74.5 & 0.54 & 42.2 \\
    \bottomrule
    \end{tabular}
\end{table}
\begin{table}[t]
\caption{Performance of LLaVA-OV and Qwen2-VL models on ActPlan-1K with the integration of TER and CAR modules, demonstrating the transferability and generalization capability of our proposed components.}
\centering
\begin{tabular}{ll@{\hspace{6pt}}c@{\hspace{6pt}}c@{\hspace{6pt}}c}
\toprule
\multirow{2}{*}{Model} & \multirow{2}{*}{Configuration} & \multicolumn{3}{c}{ActPlan-1K (total)} \\
\cmidrule(lr){3-5}
& & {\hspace{6pt}}Exec. & {\hspace{6pt}}LCS & {\hspace{6pt}}Corr. \\
\midrule
\multirow{2}{*}{LLaVA-OV \cite{li2024llava-VLM1}} & w/o TER \& CAR &
43.8 & 0.52 & 30.1\\
 & w  TER \& CAR &
\best{53.5} & \best{0.58} & \best{39.5}\\
\cline{1-5}
\multirow{2}{*}{Qwen2-VL \cite{wang2024qwen2-VLM1}} & w/o TER \& CAR & 
50.5 & 0.57 & 32.9\\
 & w  TER \& CAR & 
\best{58.3} & \best{0.59} & \best{41.1}\\
\bottomrule
\end{tabular}
\label{tab:trans}
\end{table}

As shown in Table ~\ref{tab:reranker_strategies}, the bitwise OR strategy achieved the highest accuracy among the tested options, reaching 42.0\% and 48.9\% on the two datasets respectively. In contrast, other strategies either computed mask weights too coarsely or failed to explicitly capture crucial objects through direct embedding similarity alone. This indicated that a global aggregation method is both simple and effective for consolidating multiple sentence-level masks.

\subsection{Transferability of LLaPa}

To demonstrate the transferability of our designed modules, we conducted experiments by integrating TER and CAR into two open-set multimodal models, LLaVA-OV \cite{li2024llava-VLM1} and Qwen2-VL \cite{wang2024qwen2-VLM1}. As shown in Table \ref{tab:trans}, both models exhibited performance improvements across key metrics after incorporating our proposed modules: LLaVA-OV's correctness increased from 30.1\% to 39.5\%, while Qwen2-VL advanced from 32.9\% to 41.1\% in correctness, with similar gains in executability and LCS metrics.

These results verified the general applicability and robustness of our design, highlighting its potential to enhance a wide range of multimodal models in procedural planning tasks.

\section{Conclusion}
In this paper, we present LLaPa, a novel vision-language model framework designed to address challenges in multimodal procedural planning, particularly focusing on task semantic ambiguity and counterfactual activity adaptability. By leveraging explicit task-oriented segmentation to construct task-sensitive visual representations and the Counterfactual Activities Retriever to enhance counterfactual reasoning, our experiments demonstrate that the LLaPa model generates action sequences with higher executability and the ability to handle counterfactual activities. 

\textbf{Limitations.} Despite its promising performance, LLaPa relies heavily on comprehensive textual task descriptions, which may limit its effectiveness. Additionally, the current approach does not support dynamic video context integration, which could further enhance the model's adaptability in real-world scenarios.

\section*{Acknowledgments}
This work was supported by the Aeronautical Science Foundation of China (Grant No. 2024M071077003) and the National Science Foundation for Young Scholars of China (Grant No. 62407012).
\bibliographystyle{ACM-Reference-Format}
\bibliography{bibfile}

\end{document}